\documentclass{article}

\usepackage{booktabs}
\usepackage{multirow}
\usepackage{caption}
\usepackage{placeins}
\usepackage[nonatbib, final]{neurips_2023}

\usepackage{xspace}
\usepackage[dvipsnames, table]{xcolor}

\usepackage{diagbox}

\usepackage{pifont}
\newcommand{\diffup}[1]{{\color{OliveGreen}{($\uparrow$ #1)}}}
\newcommand{\diffdown}[1]{{\color{BrickRed}{($\downarrow$ #1)}}}

\makeatletter
\DeclareRobustCommand\onedot{\futurelet\@let@token\@onedot}
\def\@onedot{\ifx\@let@token.\else.\null\fi\xspace}

\makeatother

\newcommand{\myparagraph}[1]{\vspace{0.1cm}\noindent\textbf{#1.}}

\newcommand{\coco}{COCO-Stuff\xspace}
\newcommand{\cityscapes}{Cityscapes\xspace}
\newcommand{\pascal}{PascalVOC\xspace}
\newcommand{\pascali}{PASCAL-5i\xspace}
\newcommand{\flood}{Flood\xspace}
\newcommand{\livecell}{LIVECell\xspace}
\newcommand{\shrec}{SHREC\xspace}
\newcommand{\ade}{ADE20K\xspace}
\newcommand{\dentalt}{Dental-teeth\xspace}

\newcommand{\dentalm}{Dental-jaw\xspace}
\newcommand{\indist}{In-distribution\xspace}
\newcommand{\outdist}{Out-of-distribution\xspace}

\newcommand{\best}[1]{\textbf{#1}}

\usepackage[utf8]{inputenc} 
\usepackage[T1]{fontenc}    
\usepackage{hyperref}       
\usepackage{url}            
\usepackage{booktabs}       
\usepackage{amsfonts}       
\usepackage{nicefrac}       
\usepackage{microtype}      
\usepackage{xcolor}         

\usepackage{graphicx}
\usepackage{soul}
\usepackage{amsmath}
\usepackage{booktabs}

\title{
Flexible visual prompts for in-context learning in computer vision}

\author{%
\textbf{Thomas Foster} \quad \textbf{Ioana Croitoru} \quad \textbf{Robert Dorfman} \quad \textbf{Christoffer Edlund} \\ \textbf{Thomas Varsavsky} \quad \textbf{Jon Almaz\'an} \\
V7 Labs \\
\texttt{\{thomas,ioana,robert,christoffer,tom,jon\}@v7labs.com}\\
}

\begin{document}

\maketitle

\begin{abstract}
  In this work, we address in-context learning (ICL) for the task of image segmentation, introducing a novel approach that adapts a modern Video Object Segmentation (VOS) technique for visual in-context learning. This adaptation is inspired by the VOS method's ability to efficiently and flexibly learn objects from a few examples. Through evaluations across a range of support set sizes and on diverse segmentation datasets, our method consistently surpasses existing techniques. Notably, it excels with data containing classes not encountered during training. Additionally, we propose a technique for support set selection, which involves choosing the most relevant images to include in this set. By employing support set selection, the performance increases for all tested methods without the need for additional training or prompt tuning. The code can be found at \url{https://github.com/v7labs/XMem_ICL/}.

\end{abstract}
\section{Introduction}

In-context learning (ICL) is a new paradigm primarily derived from large autoregressive language models, notably GPT-3~\cite{brown2020language}. Instead of updating model weights, ICL employs domain specific input-output pairings, termed \textit{prompts} or a \textit{support set}, which are integrated at test time to direct the model towards the intended output. This approach has proven effective in various applications with large language models (LLMs), including question-answering~\cite{li2023few,huang2023enhancing}, translation~\cite{liu2021makes,agrawal2022context} and sentiment analysis~\cite{amin2023will}. For example, at test time, a model can be guided to perform sentiment analysis by providing a structured input that pairs a statement with its sentiment label~\cite{amin2023will}. 

ICL offers the benefits of few-shot learning~\cite{lang2022learning,wang2020generalizing} without the need for weight modification or re-training when switching tasks or domains. For enterprises deploying models, ICL is desirable, allowing economical deployment of a single model that can be used for various tasks, domains, or datasets through tailored prompts, without requiring re-training. Moreover, ICL models can adapt to outdated training data or newly available data by simply using different prompts at inference.

In-context learning is increasingly explored in linguistics, but its application in computer vision remains limited. Among the few studies exploring this area, three recent ones~\cite{bar2022visual, wang2023images, wang2023seggpt} are notable. They introduce a visual prompt by merging a support set of relevant image-label pairs into a large grid. However, this 'gridding' approach has several drawbacks: it is resource-intensive, imposes limits on the maximum resolution for inference, and fixes the grid size (i.e. support set size) for all future inferences once a model is trained.

In this paper, we are the first to explore the adaptation of recent Video Object Segmentation (VOS) methods to visual ICL, aiming to overcome the limitations of existing approaches. VOS methods, aimed at binary semantic segmentation, sequentially construct a memory of the video frames and masks. Given their design to handle videos, potentially encompassing thousands of frames, VOS methods allow for the efficient utilisation of a bigger support set. Additionally, during inference, if multiple queries are to be processed using the same support set, the memory of the VOS method can be efficiently cached to avoid redundant computations. Each frame is processed on its own, ensuring that the maximum resolution of each image is not constrained by the size of the support set. Unlike existing ICL methods, the size of the support set is flexible, allowing VOS methods to adapt at inference time if more examples become available.

Furthermore, we conduct empirical evaluations to comprehensively compare existing ICL methods with VOS methods. This comparison specifically targets the task of binary segmentation, where each pixel in an image is classified as either foreground or background. This pixel-wise classification is crucial for various applications, such as medical imaging (segmenting teeth)~\cite{helli10tooth,abdi2015automatic}, disaster response (identifying flooded areas)~\cite{floodnet}, biomedical research (cell segmentation)~\cite{edlund2021livecell}, and infrastructure maintenance (segmenting potholes and cracks in roads)~\cite{thompson2022shrec}.

Our study reveals that VOS methods exhibit superior generalization to unseen classes --classes not encountered during the training process-- compared to previous techniques, while also being computationally efficient. We additionally explore and demonstrate the significant influence of support set selection on the performance of visual ICL methods. We show that including examples in the support set that are similar to the target query boosts overall performance. Our final method, which adapts a VOS approach to ICL for binary segmentation and incorporates support set selection, outperforms all other ICL methods on unseen classes.

Our main contribution can be summarized as follows: (1) We are the first to study and adapt Video Object Segmentation (VOS) methods for visual in-context learning (ICL), thus removing the restrictions of existing ICL methods on support set size and image resolution. (2) We compare VOS methods with previous visual ICL methods, revealing that our proposed VOS method performs significantly better on unseen classes and datasets. (3) Through extensive experiments, we demonstrate the importance of support set selection in enhancing visual ICL performance. By leveraging semantic visual similarity, our approach yields substantial improvements across all tested ICL methods.
\section{Related work}
\myparagraph{'Gridding' methods} 
Current in-context learning (ICL) methods~\cite{bar2022visual,wang2023images,wang2023seggpt} form a visual prompt by combining a support set of relevant prompts along with their masks into a large grid. Through pre-training a neural network to fill in gaps in this grid, they enabled the model to apply in-context learning to new tasks. Painter~\cite{wang2023images} pre-trains a vanilla vision Transformer (ViT) using a 2x2 grid of images, on a range of both segmentation (semantic, panoptic etc) and non-segmentation tasks (monocular depth estimation, image restoration etc). In a follow up work~\cite{wang2023seggpt} the authors further optimise this method for segmentation tasks, by introducing a random recolouring data augmentation at training time in an attempt to force the model to derive colouring information from the context, instead of internalising specific colours.~\cite{bar2022visual} also forms a 2x2 gridded prompt, but explores various architectures for the in-painting task, such as VQGAN~\cite{esser2021taming}, BEiT~\cite{bao2021beit} and MAE~\cite{he2022masked}. 

Larger support sets enhance model performance by providing more information. However, once trained, these models are constrained to the support set size they were initially configured with. Adjusting to a different support set size is not inherently supported and it necessitates the application of various techniques such as feature averaging. Other solutions might involve scaling to larger grid sizes which requires substantially more compute or downscaling the resolution. Unlike these models, VOS methods can dynamically adapt to any support set size at test time.

\myparagraph{VOS methods} XMem~\cite{cheng2022xmem} and STCN~\cite{cheng2021rethinking} are the most recent in a long line of Video Object Segmentation (VOS) methods that process frames and associated support masks one by one, embedding them into a "memory" that is then decoded into a new mask prediction. STCN~\cite{cheng2021rethinking} was the first to simplify this process to use just two ResNet~\cite{he2016deep} networks, one for embedding frames and masks respectively. XMem extended STCN, adding "long-term" memory for object tracking in long (1000+ frame) videos.

AOT~\cite{yang2021associating} is another approach to VOS that leverages transformer style attention over encodings of frames. The main contribution of AOT is predominantly how to track multiple objects at once, instead of producing binary masks like most VOS methods. In this work, we focus on the XMem family of VOS methods, since it represents the state of the art in binary segmentation of long videos.

\myparagraph{Other methods} UniverSeg \cite{butoi2023universeg} is a notable alternative to the gridding approach. They focus on medical imaging and adapt a UNet~\cite{ronneberger2015u} architecture to leverage a support set via repeated applications of a novel "CrossBlock" module that uses shared convolutions between the test image and each item in the support set. The model is trained on low resolution 128x128 medical imaging tasks. PerSAM ~\cite{zhang2023personalize} is notable in that it attempts to adapt the powerful SAM~\cite{kirillov2023segment} model for the ICL setting. The process begins by executing SAM inference on a single image from the support set, during which the model's internal activations and embeddings are saved. These saved states are then utilized to influence a subsequent SAM run on the test image, aiming to segment the object identified in the support set.

\myparagraph{Few-shot semantic segmentation methods}
Few-shot segmentation (FSS) methods, such as~\cite{yang2023mianet,lu2021simpler,rakelly2018conditional,Wang2020FewShotSS}, traditionally fine-tune model parameters based on a provided support set. In contrast, in-context learning does not need retraining if new examples became available. A standout benefit of in-context learning is its swift adaptability to new classes or tasks, eliminating the computational burden associated with weight updates.

\myparagraph{Support set selection} Research in the area of support set selection is limited. Some works, such as~\cite{wang2023images}, iterate over the entire meta-support set in search of the prompt yielding the highest performance. Additionally, they perform experiments where the prompt is learned through backpropagation. However, these methodologies require both training phases and access to labels of the whole meta-support set. This requirement is an important distinction from the approach proposed in our study. Our method is designed to efficiently select support sets without the need for extensive training or access to additional labels, thereby offering an effective alternative to the existing strategies.

\section{Method}
\subsection{Task Definition}
\label{sec:dataset}

\begin{figure}
    \centering
    \includegraphics[width=0.9\linewidth]{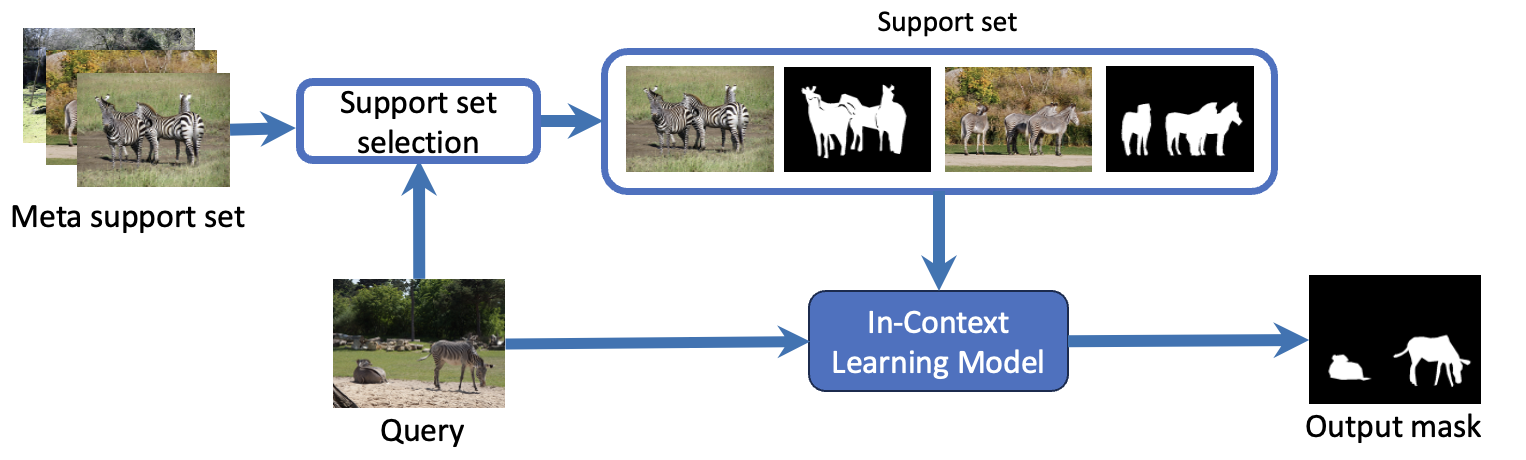}
    \caption{In-context learning for binary segmentation with support set selection. A support set is selected from a meta-support set based on similarity to the query image. This method effectively segments the query image, even in cases where it contains objects belonging to classes not seen during training.
    \label{fig:method}}
    
\end{figure}

While ICL often involves seamless task-switching, our paper narrowly focuses on proposing a method for the task of binary segmentation. The task of binary segmentation is defined as: given an image \( I \in \mathcal{R}^{H \times W \times C} \), produce a binary mask $M 
\in \{0,1\}^{H \times W}$ that identifies the pixels containing the class of interest. Despite this specialization, we continue to employ the principles of in-context learning as our proposed approach demonstrates robust generalization to unseen data. For the purposes of this study, unseen data refers to classes that the model has not encountered during any training that involves weight updates.

Given a dataset $S = ((I_1, M_1), ...(I_n, M_n))$ where $I_i$ is an image and $M_i$ a binary mask, we divide $S$ into a set $S_{\text{meta-support}}$ of meta-support pairs and an evaluation set $S_{\text{eval}}$. At inference time given an image $I_i \in S_{\text{meta-support}}$, a segmentation ICL method is a map from the meta-support set and the test image to a mask prediction:
$(I_i, S_{\text{meta-support}}) \mapsto M_{\text{pred}_i}$
\noindent We will discuss how it can be beneficial to only use a subset $S_{\text{support}}$ of $S_{\text{meta-support}}$ for each image in $S_{\text{eval}}$, which we call \textbf{support set selection}. The overview of the mapping $(I_i, S_{\text{meta-support}}) \mapsto M_{\text{pred}_i}$ using support set selection is visualized in  Figure~\ref{fig:method}.

\begin{figure}
    \centering
    \includegraphics[width=0.8\linewidth]{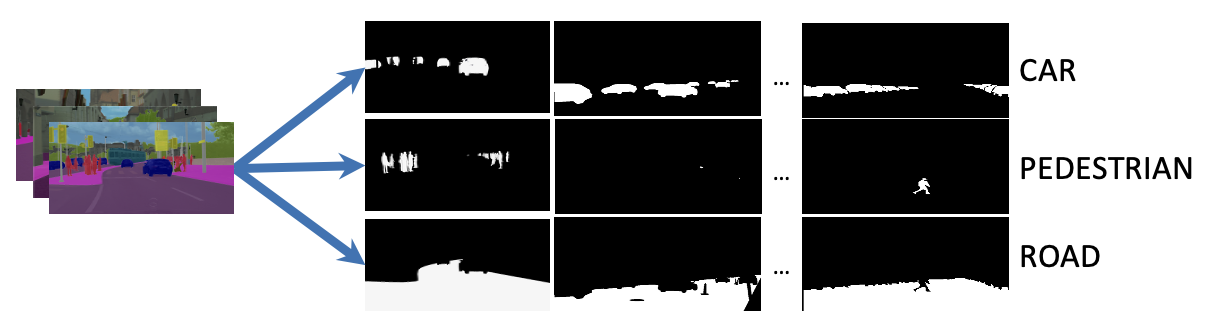}
    \caption{Construction of multiple binary segmentation datasets from one semantic segmentation dataset. Each image-mask pair in the semantic segmentation dataset is divided into multiple image-mask pairs for binary segmentation, one for each class present in the semantic segmentation.
    \label{fig:dataset}}
    
\end{figure}

\subsection{Visual ICL with XMem}
\label{sec:xmem}
As mentioned before, in this study we focus on the XMem family of
VOS methods. XMem~\cite{cheng2022xmem} employs a distinctive memory architecture with three interconnected memory stores: a frequently updated sensory memory, a high-resolution working memory, and a compact long-term memory. It consists of three end-to-end trainable networks: a \textit{query encoder} extracting features from the query image, a \textit{decoder} generating object masks from memory output, and a \textit{value encoder} merging image and mask to derive new memory features. Unlike 'gridding' in-context learning methods, XMem handles individual images and masks, aligning more with prevalent image-to-embedding approaches in vision research.

XMem begins by storing the initial embeddings of image-mask pairs in memory. When a new frame is introduced, XMem compares its embedding pixelwise to all embeddings stored in memory. This comparison results in an affinity map that highlights the regional similarities between frames. Subsequently, the decoder generates a new mask based on this affinity map. This newly created mask is embedded into the memory to be utilized for processing subsequent frames. To enhance this process further, XMem incorporates an additional long-term memory component. When the capacity of the original memory is reached, it undergoes condensation through the application of a k-nearest neighbour algorithm. For further details, we refer the reader to~\cite{cheng2022xmem}. 

To adapt XMem for in-context learning, we initially populate the model's memory with embeddings of image-mask pairs derived from the meta-support set $S_{\text{meta-support}}$. XMem inference is then run on the test image, leveraging the memory  built up from the meta-support set.

\subsection{Support set selection}
\label{sec:support}
Despite the fact that we can use up to over a thousand pairs as a support set for a VOS method, we found that segmentation performance plateaued after support sets of around size 20.
This motivated the exploration of how best to choose the support set from a large pool of labelled images to choose from. We hypothesised that having images similar to the query in the support set would help performance. To test this hypothesis, we extract image embeddings (such as CLIP~\cite{radford2021learning}) from $S_{\text{meta-support}}$. At test time, for each test image $I_{\text{test}}$, we select the support set $S_{\text{support}} \subseteq S_{\text{meta-support}}$ as the $N$ nearest neighbours to the embedding of $I_{\text{test}}$ from $S_{\text{meta-support}}$. It is important to note that unlike other methods from the literature~\cite{wang2023seggpt,wang2023images} which perform prompt tuning, our strategy for support set selection does not demand additional training. We refer to XMem models which use support set selection as \textbf{XMem+s}\pmb{$^3$}.

\subsection{Training}
Given that XMem is trained in a supervised manner on video frames -- with minimal appearance changes between consecutive frames -- we hypothesize that training the model on a dataset comprised of diverse support set images can enhance its performance. To test this hypothesis, we produced a large binary segmentation dataset. This dataset was constructed by taking semantic segmentation datasets and splitting them up into multiple binary segmentation datasets, as presented in Figure~\ref{fig:dataset}. Unlike many existing few-shot segmentation datasets such as FSS1000~\cite{FSS1000}, our approach provides numerous examples per class. We are training XMem on this dataset.

The training process begins by randomly selecting a class from a training dataset and sampling 16 image-mask pairs from its meta-support set. Treating these image-mask pairs as video frames, we conduct one training step as in the original XMem training procedure. In the end we train the VOS method on the whole dataset resulting in a single XMem finetuned model. We refer to models trained on still images as \textbf{XMem+T}.

\section{Experimental Results}
We begin by introducing the datasets used for training and evaluation in Section~\ref{sec:datasets}. This is followed by ablation studies in Section~\ref{sec:ablation} and finally we present a comparative analysis against other methods in Section~\ref{sec:comparison_sota}.

\subsection{Datasets}
\label{sec:datasets}

\myparagraph{Training datasets} For training we used three datasets, namely \ade~\cite{zhou2019semantic}, \cityscapes~\cite{cordts2016cityscapes}, \coco~\cite{caesar2018coco}. \ade~\cite{zhou2019semantic} offers a diverse array of scenes, \cityscapes~\cite{cordts2016cityscapes} focuses on urban environments, and \coco~\cite{caesar2018coco} provides a wide range of objects and complex scenes, together forming a robust foundation for training our models.

\myparagraph{Evaluation datasets} For evaluation we use 5 out-of-distribution datasets containing unseen classes that do not appear during training:
 \dentalt segmentation~\cite{helli10tooth}, \dentalm segmentation~\cite{abdi2015automatic}, \flood dataset~\cite{floodnet}, \livecell dataset~\cite{edlund2021livecell} containing cell segmentation, \shrec dataset~\cite{thompson2022shrec} containing potholes and road cracks segmentation. 
 
 Additionally, in order to assess performance on classes seen during training, experiments were conducted on the \pascali dataset~\cite{shaban2017one} which is derived from the \pascal dataset~\cite{everingham2010pascal}. It contains 20 classes, evenly divided into four folds (F1-4). It is crucial to clarify that while our final model was not directly trained on either the \pascal or \pascali datasets, classes similar to those contained in these datasets are present in the training data. Following previous methods, we report mean intersection-over-union (mIoU) metric for all experiments.

\subsection{Ablation studies}
\label{sec:ablation}

In this section, our objective is to address two important questions. Firstly, we investigate how the selection of the support set influences performance. To provide a comprehensive understanding, we present results for both \indist and \outdist datasets, offering a broad perspective on our method's effectiveness across different scenarios. Secondly, we explore the impact of finetuning XMem on image datasets. This is particularly of interest since as previously noted, XMem was originally trained on video frames, where changes in scenery are typically more gradual and less pronounced compared to still image datasets. Understanding how this transition to image datasets influences performance is therefore essential.

For both abaltion studies, the experimental results are categorized into two groups: \indist datasets and \outdist datasets. \indist refers to the evaluation splits of \cityscapes and \coco datasets, while \outdist refers to datasets that were unseen during training, specifically \dentalt, \dentalm, \flood, \livecell and \shrec. Mean results over the datasets are presented for both categories, with explicit results per each dataset available in the Appendix.
\begin{table}
\centering
\vspace{-4mm}
\caption{\textbf{Impact of support set selection (s$^3$) on \indist and \outdist datasets.} We report mIoU for different ICL methods with and without support set selection (s$^3$) using different support set sizes. + T performs image training. $^\dagger$ our proposed solution of averaging over logits.}
\begin{tabular}{@{}lccc|cc@{}}
\toprule
              &                     & \multicolumn{2}{c}{In-distribution datasets} & \multicolumn{2}{c}{Out-of-distribution datasets} \\
              \cmidrule(lr){3-4} \cmidrule(lr){5-6} 
              &                     & \multicolumn{2}{c}{Support set size} & \multicolumn{2}{c}{Support set size} \\
Method        &  s$^3$  & 1            & 10           & 1            & 10           \\
\midrule
Universeg~\cite{butoi2023universeg} & - & 5.75 & 11.89 & 22.06 & 44.84 \\
Universeg~\cite{butoi2023universeg} & \checkmark & 7.65 \diffup{1.9} & 15.51 \diffup{3.62} & 26.62 \diffup{4.56} & 51.22 \diffup{6.38} \\
PerSAM~\cite{zhang2023personalize} & - & 17.65 & - & 24.7 & - \\
PerSAM~\cite{zhang2023personalize} & \checkmark & 21.79 \diffup{4.14} & - & 31.2 \diffup{6.5} & - \\
STCN~\cite{cheng2021rethinking} & - & 11.34 & 13.28 & 52.39 & 52.92 \\
STCN~\cite{cheng2021rethinking} & \checkmark & 18.74 \diffup{7.4} & 18.62 \diffup{5.34} & 66.11 \diffup{13.71} & 56.40 \diffup{3.48} \\
SegGPT~\cite{wang2023seggpt} & - & 31.31 & 36.56 & 49.77 & 52.38 \\
SegGPT~\cite{wang2023seggpt}  & \checkmark  & 36.18 \diffup{4.97} & 38.0 \diffup{1.44} & 63.09 \diffup{13.32} & 63.17 \diffup{10.79} \\
SegGPT logit$^\dagger$ & - & 31.31 & 40.52 & 49.77 & 58.84 \\
SegGPT logit$^\dagger$ & \checkmark & 36.18 \diffup{4.87} & 43.86 \diffup{3.34} & 63.09 \diffup{13.32} & 65.12 \diffup{6.28} \\
XMem~\cite{cheng2022xmem} & - & 12.84 & 17.83 & 54.56 & 65.89 \\
XMem~\cite{cheng2022xmem} & \checkmark & 20.07 \diffup{7.23} & 25.36 \diffup{7.63} & 68.97 \diffup{14.41} & 73.34 \diffup{7.45} \\
XMem + T & - & 17.81 & 26.34 & 54.52 & 65.6 \\
XMem + T & \checkmark & 25.04 \diffup{7.23} & 32.74 \diffup{6.4} & 67.32 \diffup{12.8} & 71.83 \diffup{6.23} \\
\bottomrule
\end{tabular}
\label{tab:support_set}
\end{table}
\subsubsection{Support set selection}
We conduct support set selection experiments, presenting results for seven methods in Table~\ref{tab:support_set}. The tested methods are Universeg~\cite{butoi2023universeg}, PerSAM~\cite{zhang2023personalize}, STCN~\cite{cheng2021rethinking}, SegGPT~\cite{wang2023seggpt}, SegGPT with logits averaging, XMem and XMem+T. Each method improves with support set selection, some by over 14\%. CLIP~\cite{radford2021learning} features were used for all support set experiments.

\begin{table}
\centering
\caption{\textbf{Impact of finetuning on XMem performance}. We report mIoU on \indist and \outdist datasets and compare XMem with and without training with image datasets (+ T), for different support set sizes. The results are presented with support set selection.}
\begin{tabular}{@{}lcc|cc@{}}
\toprule
 & \multicolumn{2}{c}{\indist datasets} & \multicolumn{2}{c}{\outdist datasets} \\
 \cmidrule(lr){2-3} \cmidrule(lr){4-5}
  & \multicolumn{2}{c}{Support set size} & \multicolumn{2}{c}{Support set size} \\
Method & 1 & 10 & 1 & 10 \\
\midrule
XMem~\cite{cheng2022xmem} & 20.07 & 25.36  & 68.97  & 73.34 \\
XMem + T & 25.04 \diffup{4.96} & 32.74 \diffup{7.38}  & 67.32 \diffdown{1.65}  & 71.83 \diffdown{1.51}  \\
\bottomrule
\end{tabular}
\label{tab:finetuning}
\end{table}

\subsubsection{Training}
We explore the impact of training XMem on a large dataset of images, as detailed in Section~\ref{sec:xmem}. The results are presented in Table~\ref{tab:finetuning} on both \indist and \outdist datasets, while detailed results per dataset are presented in the Appendix. As it can be seen, training on a large dataset of images enhances performance on \indist datasets while slightly decreasing it on \outdist datasets. However, the trained model (XMem + T) represents a favorable compromise, offering improved performance on a diverse range of classes. It excels particularly on classes it has seen, making it a viable choice for those seeking a balance in performance across various datasets.

\begin{table}[h!]
\centering
\caption{\textbf{Results on \pascali}. We report results on \pascali on all folds (F0, F1, F2, F3) and compare Xmem with support set selection (s$^3$) to other ICL methods. SegGPT~\cite{wang2023seggpt} (in grey) is not comparable because it was trained on \pascal. $^\dagger$ our proposed solution of averaging over logits.\label{tab:pascal5i}}
\footnotesize 
\setlength\tabcolsep{4pt} 
\begin{tabular}{lccccc|ccccc}
\toprule
& \multicolumn{5}{c}{Support set size 1} & \multicolumn{5}{c}{Support set size 10} \\
\cmidrule(lr){2-6} \cmidrule(lr){7-11}
Method & F0 & F1 & F2 & F3 & Mean & F0 & F1 & F2 & F3 & Mean \\
\midrule

Visual prompting~\cite{bar2022visual} & 27.83 & 30.44 & 26.15 & 24.25 & 27.16 & -- & -- & -- & -- & -- \\
\textcolor{gray}{SegGPT~\cite{wang2023seggpt}}  & \textcolor{gray}{66.92} & \textcolor{gray}{73.55} & \textcolor{gray}{76.90} & \textcolor{gray}{72.44} & \textcolor{gray}{72.45} & \textcolor{gray}{68.89} & \textcolor{gray}{76.70} & \textcolor{gray}{81.47} & \textcolor{gray}{76.50} & \textcolor{gray}{75.89} \\

Universeg~\cite{butoi2023universeg}  & 12.66 & 13.53 & 12.43 & 12.60 & 12.80 & 21.52 & 22.34 & 22.14 & 22.31 & 22.08 \\

PerSAM~\cite{zhang2023personalize}  & 48.13 & 43.61 & \best{52.87} & 42.72 & 46.83 & - & - & - & - & - \\

STCN~\cite{cheng2021rethinking}  & 27.14 & 28.96 & 27.20 & 25.40 & 27.17 & 31.10 & 25.79 & 28.46 & 20.44 & 26.45 \\

\hline

\textcolor{gray}{SegGPT logits-avg$^\dagger$+s$^3$}  & \textcolor{gray}{70.09} & \textcolor{gray}{76.91} & \textcolor{gray}{78.70} & \textcolor{gray}{74.55} & \textcolor{gray}{75.06} & \textcolor{gray}{69.73} & \textcolor{gray}{78.85} & \textcolor{gray}{81.14} & \textcolor{gray}{76.56} & \textcolor{gray}{76.57}\\

XMem + s$^3$ & 47.41 & 45.49 & 43.23 & 38.19 & 43.58 & 53.82 & 52.77 & 51.60 & 41.68 & 49.97 \\

XMem + T + s$^3$ & \best{49.56} & \best{50.78} & 49.28 & \best{46.26} & \best{48.97} & \best{58.65} & \best{60.16} & \best{60.25} & \best{53.65} & \best{58.18} \\

\bottomrule
\end{tabular}
\end{table}

\subsection{Comparison with other ICL methods}
\label{sec:comparison_sota}

\begin{table}[h!]
\centering
\caption{\textbf{Generalisation results on \outdist datasets}. We report mIoU on 5 out-of-distribution datasets and compare XMem and XMem finetuned with support set selection (s$^3$) to other ICL methods using support set sizes 1 and 10. $^\dagger$ our proposed solution of averaging over logits.\label{tab:generalisation}}
\footnotesize 
\setlength\tabcolsep{4pt} 
\begin{tabular}{lcccccccccc}
\toprule
& \multicolumn{2}{c}{\dentalt} & \multicolumn{2}{c}{\dentalm} & \multicolumn{2}{c}{\flood} & \multicolumn{2}{c}{\livecell} & \multicolumn{2}{c}{\shrec} \\
\cmidrule(lr){2-3} \cmidrule(lr){4-5} \cmidrule(lr){6-7} \cmidrule(lr){8-9} \cmidrule(lr){10-11}
Method & 1 & 10 & 1 & 10 & 1 & 10 & 1 & 10 & 1 & 10 \\
\midrule
SegGPT~\cite{wang2023seggpt}   &72.76& 76.26 &74.04& \best{83.70} &38.70& 39.02 &40.58& 41.30 &22.78& 21.63 \\ 

Universeg~\cite{butoi2023universeg}  &40.67& 67.23 &42.20& 73.46 &8.77& 21.45 &15.53& 41.97 &3.11& 20.07 \\ 

PerSAM~\cite{zhang2023personalize}    &21.37& - &47.73& - &20.18& - &24.06& - &10.14& - \\ 

STCN~\cite{cheng2021rethinking} & 77.35 & 67.72 & 73.10 & 67.96 & 34.30 & 41.96 & 54.39 & 57.74 & 22.82 & 29.21 \\

\hline

SegGPT logits-avg$^\dagger$ + s$^3$   &74.53& 77.51 &79.54& 69.15 &66.05& 64.81 &59.94& 70.45 &35.41& 43.67 \\ 
XMem + s$^3$  &\best{79.93}& \best{82.88} &\best{80.75}& 83.58 &\best{67.54}& \best{72.82} &\best{78.68}& \best{80.57} &37.93& \best{46.84} \\ 

XMem + T + s$^3$  &76.87& 80.97 &77.93& 83.13 &67.24& 72.50 &76.11& 77.90 &\best{38.45}& 44.67 \\ 
\bottomrule
\end{tabular}
\end{table}

We evaluated our final method (XMem + T + support set selection) against several methods, namely: Visual prompting, Universeg~\cite{butoi2023universeg}, PerSAM~\cite{zhang2023personalize}, STCN~\cite{cheng2021rethinking}, and SegGPT~\cite{wang2023seggpt}. It is important to highlight a significant observation made during our experiments: SegGPT method typically averages features when the support set contains more than one example, while our experiments revealed a superior approach. Specifically, we found that processing one example at a time and averaging logits across the entire support set yields improved results — sometimes enhancing performance by over 10\% — especially for \outdist datasets (as shown in Table~\ref{tab:generalisation}). Further, results are presented using both the original SegGPT and our modified version that implements logit averaging. We start by showing results on Pascal-5i in Table~\ref{tab:pascal5i}.

\begin{figure}[h!]
    \centering
    \includegraphics[width=\linewidth]{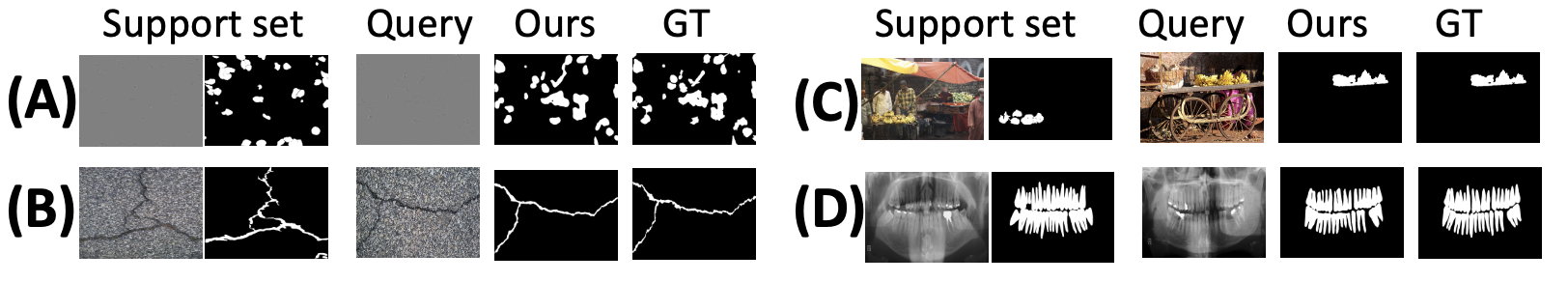}
    \caption{Qualitative results on (A) \livecell~\cite{edlund2021livecell}, (B) \shrec~\cite{thompson2022shrec}, (C) \coco~\cite{caesar2018coco}, (D) \dentalt~\cite{helli10tooth}. Ours represents XMem with support set selection. The same model weights are used to segment a variety of classes, such as biological cells, bananas, dental radiography and road surface cracks. 
    \label{fig:qualitative}}
\end{figure}

\myparagraph{Generalisation Results} We present results on five out-of-distribution datasets in Table~\ref{tab:generalisation}. XMem with support set selection shows superior generalisation capabilities, notably improving performance on the \flood and \livecell datasets. 

\noindent\textbf{Qualitative results} In Figure~\ref{fig:qualitative} we present qualitative results on four datasets. We refer to XMem with support set selection as ours. We observe that by using the same model, specifically XMem + s$^3$, we are able to segment various classes.
\section{Conclusion}
Our study targets visual in-context learning (ICL) specifically for binary semantic segmentation tasks. We are the first to explore the application of Video Object Segmentation (VOS) methodologies to in-context learning.
Through extensive analysis, we demonstrate that our adapted VOS methodology offers superior generalization to unseen classes, providing a more robust and versatile solution than existing ICL methods.

In addition, we also explore the significance of support set selection in enhancing the performance of visual in-context learning (ICL). We found that a strategic approach to choosing support sets can lead to considerable improvements in performance. This becomes particularly notable as we observed that such selection of support sets yielded enhanced results across all seven ICL methods we tested. 
\FloatBarrier

{\small
\bibliographystyle{plain}
\bibliography{egbib}
}
\clearpage
\appendix

\section{Datasets}
In Table~\ref{tab:dataset-description} we show how many images are present in the training and evaluation for each of the datasets used throughout the paper. 
As specified in the main paper for training we used:
\ade~\cite{zhou2019semantic}, \cityscapes~\cite{cordts2016cityscapes}, \coco~\cite{caesar2018coco}. We also conducted assessments on the evaluation set of those datasets, with results presented in Table~\ref{tab:training_datasets_appx}. Additional evaluations were performed on  \dentalt segmentation~\cite{helli10tooth}, \dentalm segmentation~\cite{abdi2015automatic}, \flood dataset~\cite{floodnet}, \livecell dataset~\cite{edlund2021livecell} containing cell segmentation, \shrec dataset~\cite{thompson2022shrec} containing potholes and road cracks segmentation and on \pascali dataset.

\begin{table}[h]
  \centering
  \caption{\textbf{Datasets used for training and evaluation.} $^*$ \pascali dataset has 20 classes split equally in 4 folds.}
  \begin{tabular}{lllll}
    \toprule
                     & \multicolumn{2}{c}{Training} & \multicolumn{2}{c}{Evaluation} \\
    \cmidrule(lr){2-3} \cmidrule(lr){4-5}
    Dataset name     & \#classes & \#pairs & \#classes & \#pairs\\
    \midrule
    \ade & 130 & 165120  & 0 & 0  \\
    \coco & 171 & 965042 & 171 & 42095     \\
    \cityscapes     & 19 & 34723 & 19 & 6006 \\
    \midrule
    \pascali & 0 & 0 & 20$^*$ & 1964 \\
    \dentalt & 0 & 0 & 1 & 47 \\
    \dentalm & 0 & 0 & 2 & 47 \\
    \flood & 0 & 0 & 9 & 1740 \\
    \livecell & 0 & 0 & 1 & 1512 \\
    \shrec & 0 & 0 & 3 & 290 \\
    
    \bottomrule
\end{tabular}
  \label{tab:dataset-description}
\end{table}

 \dentalt segmentation~\cite{helli10tooth}, \dentalm segmentation~\cite{abdi2015automatic}, \flood dataset~\cite{floodnet}, \livecell dataset~\cite{edlund2021livecell} containing cell segmentation, \shrec dataset~\cite{thompson2022shrec} containing potholes and road cracks segmentation. 

\section{Additional Results}
Below we present the detailed results of all the methods and on all datasets. In Tab.~\ref{tab:pascal5i_appx} we present the results of all tested methods on \pascali. In Table~\ref{tab:training_datasets_appx} we present results of all methods on the evaluation split of all training datasets. Further, we present additional results in Table~\ref{tab:generalisation_appx} for \outdist datasets.

\begin{table}[h]
\caption{\textbf{Results on \pascali}. We report results on \pascali on all folds (F0, F1, F2, F3) and compare Xmem to other ICL methods. We extend Table~\ref{tab:pascal5i} and present here all variants for support set selection (s$^3$). SegGPT~\cite{wang2023seggpt} (in grey) is not comparable because it was trained on \pascal. $^\dagger$ our proposed solution of averaging over logits.\label{tab:pascal5i_appx}}
\centering
\footnotesize 
\setlength\tabcolsep{4pt} 
\begin{tabular}{llccccc|ccccc}
\toprule
& & \multicolumn{5}{c}{Support set size 1} & \multicolumn{5}{c}{Support set size 10} \\
\cmidrule(lr){3-7} \cmidrule(lr){8-12}
Method & $s^3$ & F0 & F1 & F2 & F3 & Mean & F0 & F1 & F2 & F3 & Mean \\
\midrule
Visual prompting~\cite{bar2022visual} & - & 27.83 & 30.44 & 26.15 & 24.25 & 27.16 & -- & -- & -- & -- & -- \\
\textcolor{gray}{SegGPT~\cite{wang2023seggpt}} & - & \textcolor{gray}{66.92} & \textcolor{gray}{73.55} & \textcolor{gray}{76.90} & \textcolor{gray}{72.44} & \textcolor{gray}{72.45} & \textcolor{gray}{68.89} & \textcolor{gray}{76.70} & \textcolor{gray}{81.47} & \textcolor{gray}{76.50} & \textcolor{gray}{75.89} \\

\textcolor{gray}{SegGPT~\cite{wang2023seggpt}} & \checkmark & \textcolor{gray}{70.09} & \textcolor{gray}{76.91} & \textcolor{gray}{78.70} & \textcolor{gray}{74.55} & \textcolor{gray}{75.06} & 
\textcolor{gray}{71.22} & \textcolor{gray}{79.26} & \textcolor{gray}{81.88} & \textcolor{gray}{77.02} & \textcolor{gray}{77.34} \\

Universeg~\cite{butoi2023universeg} & - & 12.66 & 13.53 & 12.43 & 12.60 & 12.80 & 21.52 & 22.34 & 22.14 & 22.31 & 22.08 \\
Universeg~\cite{butoi2023universeg} & \checkmark & 17.54 & 16.97 & 16.50 & 16.97 & 16.99 & 32.15 & 32.97 & 30.65 & 29.06 & 31.21 \\
PerSAM~\cite{zhang2023personalize} & - & 48.13 & 43.61 & \best{52.87} & 42.72 & 46.83 & - & - & - & - & - \\
PerSAM~\cite{zhang2023personalize} & \checkmark & 59.61 & 56.24 & 55.41 & 49.44 & 55.17 & - & - & - & - & - \\
STCN~\cite{cheng2021rethinking} & - & 27.14 & 28.96 & 27.20 & 25.40 & 27.17 & 31.10 & 25.79 & 28.46 & 20.44 & 26.45 \\
STCN~\cite{cheng2021rethinking} & \checkmark & 44.64 & 43.32 & 42.40 & 37.13 & 41.87 & 40.25 & 32.54 & 35.38 & 22.42 & 32.65 \\
\hline
\textcolor{gray}{SegGPT logits-avg} & - & \textcolor{gray}{66.92} & \textcolor{gray}{73.44} & \textcolor{gray}{76.24} & \textcolor{gray}{73.30} & \textcolor{gray}{72.65} & \textcolor{gray}{68.43} & \textcolor{gray}{78.42} & \textcolor{gray}{80.65} & \textcolor{gray}{77.30} & \textcolor{gray}{76.46} \\

\textcolor{gray}{SegGPT logits-avg$^\dagger$} & \checkmark & \textcolor{gray}{70.09} & \textcolor{gray}{76.91} & \textcolor{gray}{78.70} & \textcolor{gray}{74.55} & \textcolor{gray}{75.06} & \textcolor{gray}{69.73} & \textcolor{gray}{78.85} & \textcolor{gray}{81.14} & \textcolor{gray}{76.56} & \textcolor{gray}{76.57}\\

XMem  & - & 34.10 & 32.34 & 29.05 & 28.52 & 31.00 & 42.57 & 41.71 & 38.59 & 33.55 & 39.10 \\
XMem & \checkmark & 47.41 & 45.49 & 43.23 & 38.19 & 43.58 & 53.82 & 52.77 & 51.60 & 41.68 & 49.97 \\

XMem + FT & - & 34.20 & 38.04 & 34.34 & 34.28 & 35.22 & 47.23 & 50.82 & 49.87 & 46.30 & 48.56\\

XMem + FT & \checkmark & \best{49.56} & \best{50.78} & 49.28 & \best{46.26} & \best{48.97} & \best{58.65} & \best{60.16} & \best{60.25} & \best{53.65} & \best{58.18} \\
\bottomrule
\end{tabular}
\end{table}

\begin{table}[h]
\centering
\caption{\textbf{Results on \cityscapes, and \coco.} We report mIoU for different ICL methods on the evaluation splits of the training datasets, with and without support set selection (s$^3$), and using different support set sizes. + T performs training on image datasets. $^\dagger$ averaging over logits and support set selection. \label{tab:training_datasets_appx}}
\footnotesize 
\setlength\tabcolsep{4pt} 
\begin{tabular}{lccccc}
\toprule
& & \multicolumn{2}{c}{\cityscapes} & \multicolumn{2}{c}{\coco} \\
\cmidrule(lr){3-4} \cmidrule(lr){5-6} 
& & \multicolumn{2}{c}{Support set size} & \multicolumn{2}{c}{Support set size}  \\
Method & $s^3$ & 1 & 10 & 1 & 10  \\
\midrule
SegGPT~\cite{wang2023seggpt} & - & 35.43 & 37.27 & 27.19 & 35.84 \\
SegGPT~\cite{wang2023seggpt} & \checkmark & 36.74 &37.79  & 35.62 & 38.21 \\
Universeg~\cite{butoi2023universeg} & - & 6.45 & 13.83 & 5.05 & 9.94 \\
Universeg~\cite{butoi2023universeg} & \checkmark & 7.50 &  15.64 & 7.80 & 15.38 \\
PerSAM~\cite{zhang2023personalize} & - & 16.04 & - & 19.25 & - \\
PerSAM~\cite{zhang2023personalize} & \checkmark & 16.64 & -& 26.93 & - \\
STCN~\cite{cheng2021rethinking} & - &  13.03 & 14.53 & 9.65 & 12.03 \\
STCN~\cite{cheng2021rethinking} & \checkmark & 18.17 & 21.40 & 19.30 & 15.84 \\
\hline
segGPT logit-avg$^\dagger$ & - & 35.43 & 41.88 & 27.19 & 39.16  \\
SegGPT logit-avg$^\dagger$ & \checkmark & 36.74 & 43.31 & 35.62 & 44.41  \\
XMem & - & 15.80 & 21.94 & 9.87 & 13.71 \\
XMem & \checkmark &  20.31 & 26.37 & 19.83 & 24.55\\
XMem + FT & - &  21.51 & 30.81 & 14.10 & 21.86\\
XMem + FT & \checkmark &  25.00 & 33.59 & 25.08 & 31.88\\
\bottomrule
\end{tabular}
\end{table}

\begin{table}[h]
\centering
\caption{\textbf{Generalisation results on \outdist datasets}. We report mIoU on 5 out-of-distribution datasets and compare XMem and XMem image training (+ T) to other ICL methods using different support set sizes. We extend Table~\ref{tab:generalisation} and present here all variants for support set selection (s$^3$).  $^\dagger$ our proposed solution of averaging over logits. \label{tab:generalisation_appx}}
\footnotesize 
\setlength\tabcolsep{4pt} 
\begin{tabular}{lc|cccccccccc}
\toprule
& & \multicolumn{2}{c}{\dentalt} & \multicolumn{2}{c}{\dentalm} & \multicolumn{2}{c}{\flood} & \multicolumn{2}{c}{\livecell} & \multicolumn{2}{c}{\shrec} \\
\cmidrule(lr){3-4} \cmidrule(lr){5-6} \cmidrule(lr){7-8} \cmidrule(lr){9-10} \cmidrule(lr){11-12}
Method & $s^3$ & 1 & 10 & 1 & 10 & 1 & 10 & 1 & 10 & 1 & 10 \\
\midrule
SegGPT~\cite{wang2023seggpt} &-  &72.76& 76.26 &74.04& 83.70 &38.70& 39.02 &40.58& 41.30 &22.78& 21.63 \\ 
SegGPT~\cite{wang2023seggpt} & \checkmark &74.53& 76.02 &79.54& 83.53 &66.05& 63.84 &59.94& 55.57 &35.41& 36.88\\ 
Universeg~\cite{butoi2023universeg} & -  &40.67& 67.23 &42.20& 73.46 &8.77& 21.45 &15.53& 41.97 &3.11& 20.07 \\ 
Universeg~\cite{butoi2023universeg} & \checkmark  &40.17& 69.44 &44.70& 75.42 &20.26& 37.33 &22.95& 51.80 &5.03& 22.12 \\ 

PerSAM~\cite{zhang2023personalize} & -  &21.37& - &47.73& - &20.18& - &24.06& - &10.14& - \\ 
PerSAM~\cite{zhang2023personalize} & \checkmark & 22.19 & - & 46.81 & - & 40.78 & - & 25.85 & - & 18.22 & - \\ 
STCN~\cite{cheng2021rethinking} & - & 77.35 & 67.72 & 73.10 & 67.96 & 34.30 & 41.96 & 54.39 & 57.74 & 22.82 & 29.21 \\
STCN~\cite{cheng2021rethinking} & \checkmark & 78.67 & 66.64 & 79.44 & 67.45 & 67.37 & 58.16 & 72.55 & 55.59 & 32.50 & 34.16 \\

\hline
SegGPT logits-avg$^\dagger$ & -  &72.76& 77.33 &74.04& 66.28 &38.70& 56.28 &40.58& 54.87 &22.78& 39.46 \\ 
SegGPT logits-avg$^\dagger$ & \checkmark  &74.53& 77.51 &79.54& 69.15 &66.05& 64.81 &59.94& 70.45 &35.41& 43.67 \\ 
XMem & -  &78.51& 82.88 &73.93& \best{85.19} &32.50& 48.30 &65.14& 75.79 &22.72& 37.27\\ 
XMem & \checkmark  &\best{79.93}& \best{82.88} &\best{80.75}& 83.58 &\best{67.54}& \best{72.82} &\best{78.68}& \best{80.57} &37.93& \best{46.84} \\ 
XMem + FT & -  &75.94& 80.65 &70.72& 84.11 &36.83& 52.08 &60.39& 71.85 &28.71& 39.30 \\ 
XMem + FT & \checkmark  &76.87& 80.97 &77.93& 83.13 &67.24& 72.50 &76.11& 77.90 &\best{38.45}& 44.67 \\ 
\bottomrule
\end{tabular}
\end{table}

\end{document}